\documentclass[letterpaper, 10 pt, conference]{ieeeconf}  

\IEEEoverridecommandlockouts                              

\usepackage{amsmath} 
\usepackage{amssymb}  
\usepackage[linesnumbered,ruled]{algorithm2e}
\usepackage{gensymb}
\usepackage{graphicx}
\usepackage{epstopdf}
\usepackage{subcaption}
\usepackage{pifont}
\usepackage[colorlinks]{hyperref}
\usepackage{multirow}
\usepackage[T1]{fontenc}
\usepackage[super]{nth}
\usepackage{fix-cm}
\usepackage{floatrow}
\usepackage[]{caption}
\usepackage{lipsum}  
\DeclareMathOperator*{\argmin}{arg\,min} 

\makeatletter
\let\NAT@parse\undefined
\makeatother

\usepackage[square,comma,numbers,sort&compress]{natbib} 

\title{\LARGE \bf
TRADE: Object Tracking with 3D Trajectory and Ground Depth Estimates for UAVs
}

\author{Pedro F. Proen\c{c}a$^*$, Patrick Spieler$^*$, Robert A. Hewitt$^*$, Jeff Delaune$^*$
\thanks{$^{*}$ Jet Propulsion Laboratory, California Institute of Technology, \quad Pasadena, CA, USA}%
}

\begin{document}

\maketitle
\thispagestyle{empty}
\pagestyle{empty}

\begin{abstract}
We propose TRADE for robust tracking and 3D localization of a moving target in cluttered environments, from UAVs equipped with a single camera. Ultimately TRADE enables 3d-aware target following. 

Tracking-by-detection approaches are vulnerable to target switching, especially between similar objects. Thus, TRADE predicts and incorporates the target 3D trajectory to select the right target from the tracker's response map. Unlike static environments, depth estimation of a moving target from a single camera is a ill-posed problem. Therefore we propose a novel 3D localization method for ground targets on complex terrain. It reasons about scene geometry by combining ground plane segmentation, depth-from-motion and single-image depth estimation. The benefits of using TRADE are demonstrated as tracking robustness and depth accuracy on several dynamic scenes simulated in this work. Additionally, we demonstrate autonomous target following using a thermal camera by running TRADE on a quadcopter's board computer.

\par

\end{abstract}


\section{Introduction}

Object tracking and 3D localization from a UAV has several applications (e.g. defense, disaster response, wildlife monitoring) involving target following, which already have some commercial solutions for consumer drones (e.g. Skydio, DJI) relying on GPS beacons, stereo cameras and visual object trackers. However persistent tracking and 3D localization of a non-cooperative target from a single camera remains a challenging problem.

\begin{figure}[t]
	\centering
	\begin{tabular}{@{}c@{\hspace{2pt}}c@{}}
	\includegraphics[width=0.495\textwidth]{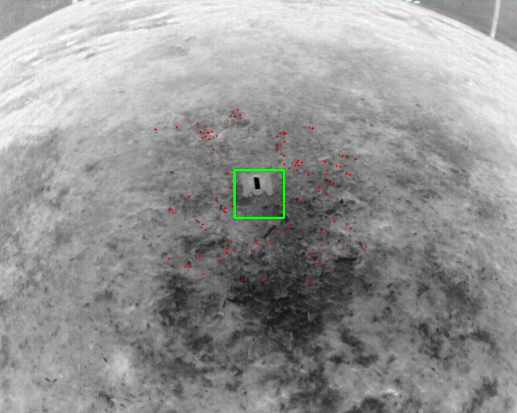} &
    \includegraphics[width=0.495\textwidth,trim={0.01cm 0cm 0cm 0cm},clip]{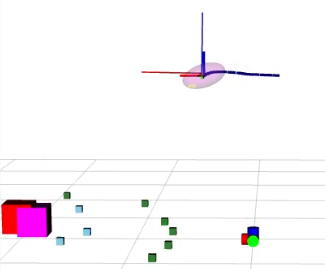} \\
    \includegraphics[width=0.495\textwidth,,trim={0.0cm 1.5cm 0cm 0.5cm},clip]{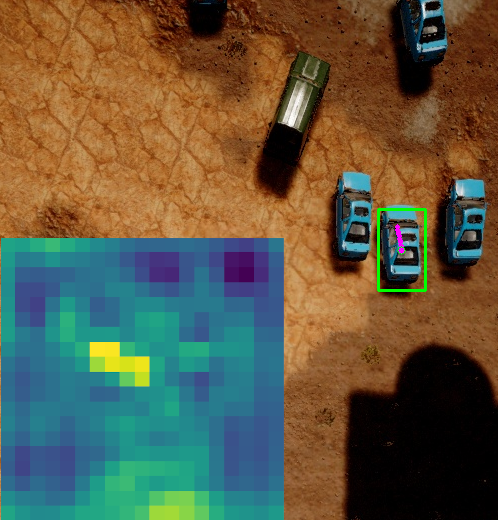} &
    \includegraphics[width=0.495\textwidth,trim={0.8cm 2.5cm 1.5cm 4.85cm},clip]{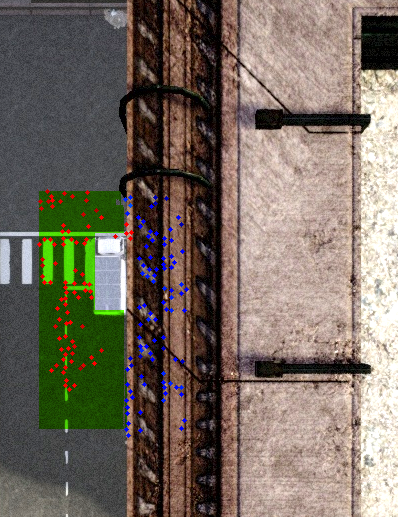} \\
    \end{tabular}
	\caption{ \textit{Top}: Drone following autonomously a remote-controlled car using a thermal camera with onboard TRADE {\footnotesize\url{https://youtu.be/QUzky1LFqpY}}. \textit{Bottom:} Two scenes from our synthetic dataset showing the benefit of our approach. \textit{Bottom-Left}: The tracker's correlation filter response \cite{bhat2019learning} shows two similar peaks, which lead to \textit{target switching} between cars. Our approach uses the trajectory predictions (seen in magenta) to select the right peak. \textit{Bottom-Right:} UAV tracking a truck from a rooftop. Our ground plane segmentation (shown in green) allows us to reject features from the building top to estimate the correct ground plane. Tracked features are color-coded based on depth. For more details refer to: {\footnotesize\url{https://youtu.be/MGPK65gm9GI}}}
	\label{fig:fig0}
\end{figure}

In terms of persistent tracking, \textit{Tracking-by-detection} has become the dominant paradigm \cite{bhat2019learning,kristan2020eighth} in generic visual object tracking thanks to learned discriminative and efficient models. Despite its success, this approach alone leads to \textit{target switching} especially between similar objects (as shown in Fig. \ref{fig:fig0}). This is mainly due to the absence of an object motion model. In this work, we propose to predict the 3D trajectory of a ground object with visual-inertial odometry (VIO) to prevent \textit{target switching} by selecting the peak from the tracker's correlation filter response map closer to the predicted location. \par

Unlike static environments, estimating the depth of a moving object from a single camera is a ill-posed problem without knowledge of the object's motion. Thus we propose a solution for ground targets by combining single-image depth estimation with depth estimates from camera motion.
While the former is used to obtain dense depth of the moving target relative to its surroundings, the latter provides sparse accurate depth measurements from the terrain. Our localization method can then infer the ground plane from the scene geometry (i.e. ground plane segmentation) to raycast the object's depth. Moreover, a temporal plane fusion step is proposed to account for temporally covered or textureless ground and missing depth-from-motion due to hovering.

Our contributions are the following:

\begin{itemize}
\item A novel 3D localization method for a dynamic ground object that is robust to high terrain relief.
\item Couple object 3D trajectory forecasting, and camera pose with a Discriminative Correlation Filter tracker to avoid target switching.
\item A photorealistic tracking dataset for UAV with ground-truth depth and poses and an extensive evaluation.
\item Demonstrate real-time UAV target following using TRADE onboard.
\end{itemize}

\section{Related Work}
Recently, in generic visual object tracking, several learning-based efforts have been made to address the problem of \textit{target switching} \cite{bhat2020know,mayer2021learning}. In Multi-Object Tracking \cite{Wojke2017simple,bergmann2019tracking}, this problem is formulated as a data association problem which is commonly addressed by using a constant-velocity Kalman Filter and a Hungarian algorithm. However this Kalman Filter works on the image space, where optical flow is non-linear and it assumes a static camera. A camera motion compensation is used in \cite{bergmann2019tracking} based on image registration and in \cite{zhang2021jointly} based on homography warping, which assumes the homography is estimated using the object's ground plane. In \cite{henein2020dynamic}, target 3D trajectory is modeled in a SLAM factor graph but it relies on a stereo-camera. In \cite{amirian2019social,coskun2017long} trajectory models are learned for human motion using LSTMs. \par

Object 3D localization from a UAV has been addressed using GPS receivers \cite{zhao2019detection}, laser range finder \cite{wang2017real}, georeferenced topographic maps \cite{dobrokhodov2006} and flat earth assumption \cite{barber2006vision}. There has been extensive work \cite{zhang2019eye,yuan20063d, liu2021moving} using ground plane estimation for 3D object localization. The most similar to our work is \cite{zhang2019eye}, which uses depth estimates from Visual Odometry and a barometer to estimate the plane normals and height but this also assumes the scene is planar. In terms of monocular object 3D localization from the ground, \cite{mousavian20173d} proposes to estimate 3D car poses by combining 2D bounding boxes, orientation regression and the object dimensions. Single-image depth networks \cite{ranftl2020towards,godard2019digging} have been demonstrating compelling results on several datasets (e.g KITTI). In this work, we investigate how these generalize to aerial downward-looking cameras.
\par

\begin{figure}[t]
	\centering
    \includegraphics[width=0.8\textwidth,trim={0cm 0cm 0cm 0cm},clip]{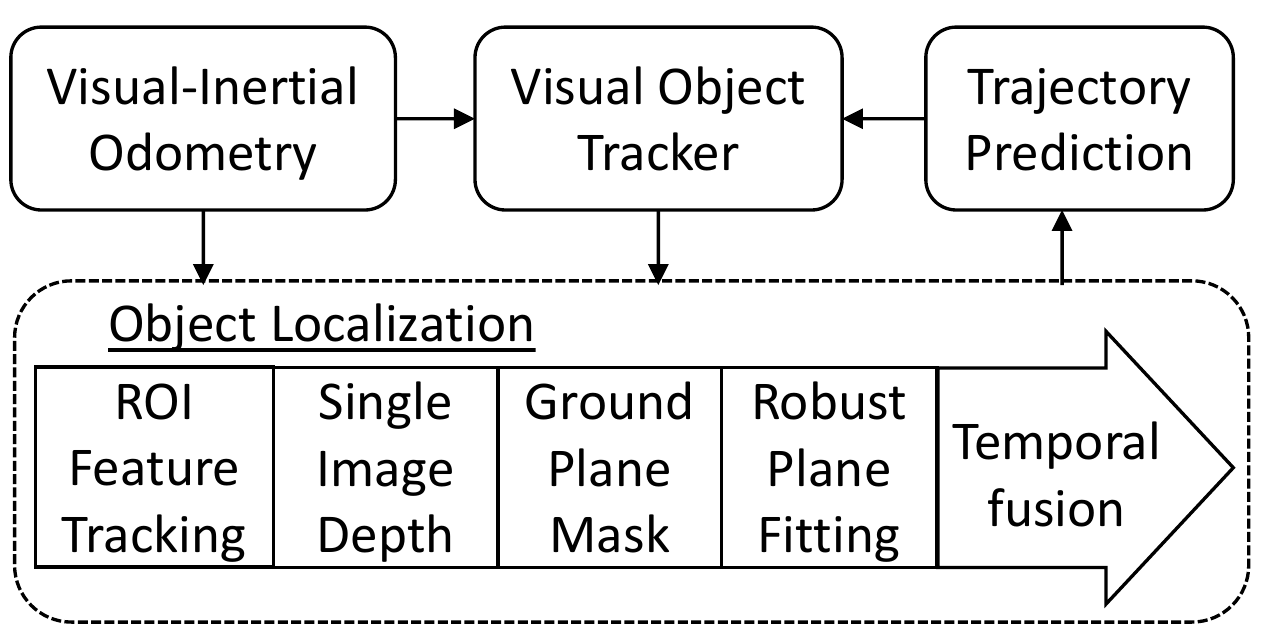}
	\caption{Overview of our object tracking and 3D localization system.}
	\label{fig:fig1}
\end{figure}

\section{System Overview}
\label{sec:overview}

Our system pipeline is shown in Fig. \ref{fig:fig1}. Firstly, a Discriminative Correlation Filter (DCF) tracker is initialized as usual with a bounding box on an initial frame. The bounding box is then used to initialize the ROI Feature Tracking by detecting Harris corners within a Region of Interest (ROI) surrounding the bounding box, which is excluded from the ROI. The ROI is then shifted as the bounding box moves through tracking. Using this dedicated feature tracking module allows to maintain a dense distribution of features around the object, without adding overhead to the VIO. \par
For every frame, depth is estimated and refined for the ROI tracks given the camera poses from VIO. These tracks can be backprojected to a point cloud and fit a plane. However, since not all tracks are from the object's ground plane, we first select tracks based on our ground plane segmentation which relies on a single image depth model to provide dense depth for both the target and the ROI. However since this is only relative depth, we use the ROI feature depth estimates to effectively scale it.

Given the resulting ground plane mask, the selected ROI tracks with 3D coordinates are used in a RANSAC multi-plane fitting routine. Since the ground plane segmentation can fail and the ROI features may not be enough, we use a temporally-fused plane model, which aggregates the inlier points from the last RANSAC plane fitting in a buffer together with inliers from past frames. The temporal fusion also includes a gating strategy to enforce temporal consistency. The aggregated points are used in a RANSAC multi-plane fitting loop once again to estimate the final plane. Then, given the target image coordinates from the DCF tracker and the camera pose we can raycast the 3D location. This is then used to update the trajectory model, whose predictions for the next frames are used to guide the DCF Tracker, as described in the next section. The remaining sections provide more details for each module of our pipeline.


\section{Tracking with trajectory estimates}
\label{sec:traj_prediction}
Visual object trackers generally output a 2D score map (shown in Fig. \ref{fig:fig0}) that maps to locations in an image search window around the previous target location. Then the location with highest score is simply selected as the new target location. \par
Instead, we first center the search window around the location predicted by our trajectory model which is projected to the current image using the camera pose. We then perform peak selection on the score map: First we normalize the score map with a softmax function, then, using Non-Maximum suppression, we select as location candidates the peaks within a certain fraction of the maximum peak and take the peak that is closer to the search window origin. \par

As a trajectory model, We use a linear Kalman filter to estimate the state $\{p,v,a\}$, respectively the object absolute 3D location, velocity and acceleration.
To prevent unbounded motion during temporary tracking loss, we use a damping factor both in velocity and acceleration instead of a constant model in the state transition. The state is updated using only the 3D location observation residuals. The process and measurement noise was set empirically.

\section{Robust Object 3D Localization}

\begin{figure}[tb]
	\centering
	\begin{tabular}{@{}c@{ }c@{}}
    \includegraphics[width=0.4\textwidth,trim={0cm 0.3cm 0cm 0cm},clip]{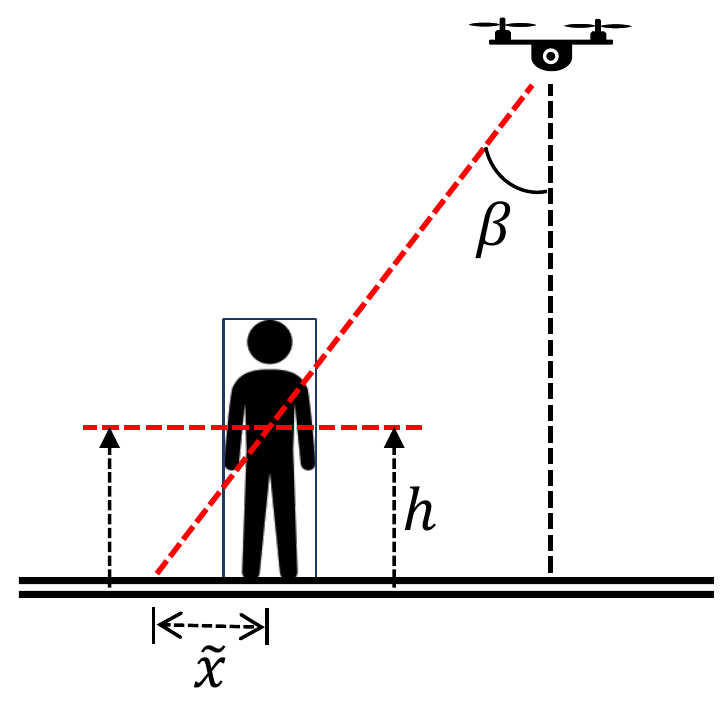} &
    \includegraphics[width=0.35\textwidth,angle=90,origin=c]{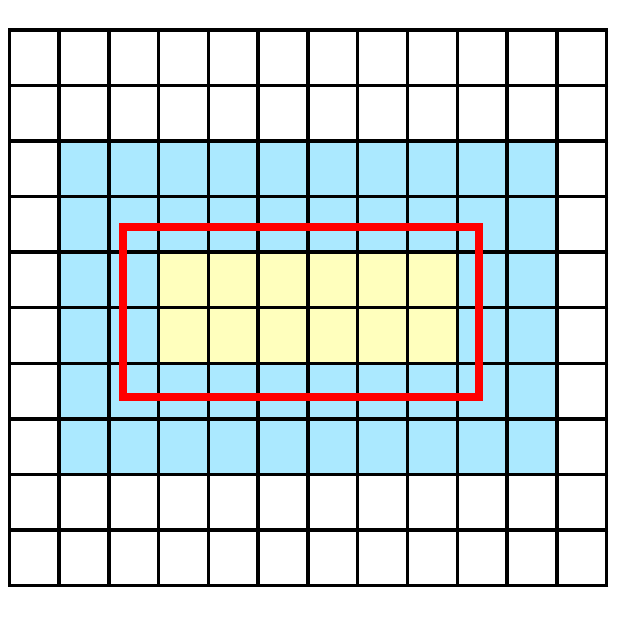} \\
    (a) & (b)

    \end{tabular}
	\caption{(a) Lateral error due to off-nadir. To minimize this error we lift the ground plane before raycasting. (b) Image grid used for segmenting the ground plane from the cropped Single-Image depth map. The red square corresponds to the target bounding box. Points within the yellow tiles are used to estimate the object height. Planes are fit to blue and white tiles. Based on the estimated planes, seeds are sampled from the blue section for Region Growing, which is performed on the white tiles.}
	\label{fig:fig2}
\end{figure}

Our object localization is based on the projection of the object bounding box center on the ground plane. However as illustrated in Fig. \ref{fig:fig2}.a, camera off-nadir $\beta$ leads to a lateral error $\tilde {x}=h \tan \beta$ where $h$ is the height at which the ray intersects the object. To reduce this error we lift the ground plane by an estimate of half the object height before raycasting the depth. The next subsections cover all modules of our localization approach.

\subsection{ROI Feature tracking}
To track a dense distribution of static features around the target, we detect Harris corners from a tiled ROI surrounding the bounding box. For simplicity, The ROI is split by a 3$\times$3 grid where the center tile corresponds to the object bounding box. For every tile (except the middle one), we maintain $N$ tracks by adding more features if necessary. These are tracked frame-to-frame using KLT. For each track, we keep the first observation and respective camera pose and triangulate depth with the current camera pose using a linear solution. These stereo depth estimates are then fused using well-established recursive multiplication of Gaussians \cite{kweon1989terrain} and initialized as 3D points once the its uncertainty drops below a certain threshold.

\subsection{Single-Image Depth}
\label{sec:SIdepth}

We use MiDaS \cite{ranftl2020towards} with the authors supplied weights as our single-image depth model because it was trained across a large mixture of different datasets and it performed the best in our preliminary results. However MiDaS only generates an affine transformation of the world-scaled inverse depth . Therefore, we estimate the translation and scale parameters $\{\theta_0,\theta_1\}$ by minimizing
\begin{equation}
\{\theta_0,\theta_1\}^* = \argmin_{\{\theta_0,\theta_1\}}\sum^{M}_{i}{\left(\theta_0+\rho^\prime_i\theta_1 -\rho_i\right)^2}
\end{equation}
in a linear least squares where $\{\rho_1, ... ,\rho_M\}$ are ROI feature inverse depth values and $\{\rho^\prime_1, ... ,\rho^\prime_M\}$ are the corresponding MiDaS map pixel values. Since the ROI depth values and MiDas may contain outliers we also minimize this in a RANSAC with a minimum set of 2 point correspondences. It is worth noting that the solution is degenerative on planes parallel to the image plane. Thus in the RANSAC sampling we enforce a minimum distance between ROI depth samples.

\begin{figure}[t]
	\centering
	\begin{tabular}{@{}c@{ }c@{}}
    \includegraphics[width=0.5\textwidth]{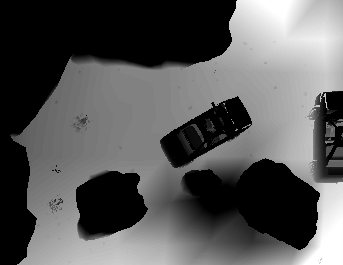} &
    \includegraphics[width=0.49\textwidth]{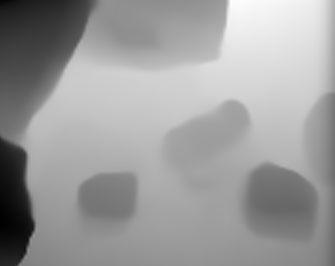} \\
    \includegraphics[width=0.5\textwidth]{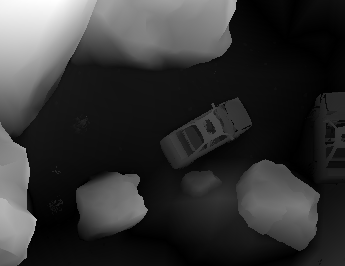} &
    \includegraphics[width=0.49\textwidth]{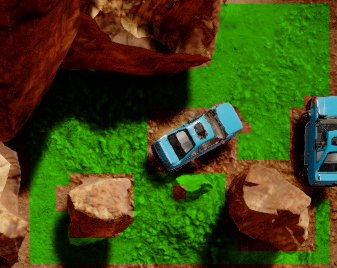}
    \end{tabular}
	\caption{\textit{Top-left:} Ground-truth depth normalized with the Single-Image depth shown on \textit{top-right}, after affine transformation. \textit{Bottom-left:} Single-Image depth error. \textit{Bottom-right:} Obtained ground mask after refinement shown in green.}
	\label{fig:fig3}
\end{figure}

One might be tempted to sample directly depth on the target from the recovered world-scaled depth map. However the error is unbounded inside the bounding box, as shown in Fig. \ref{fig:fig3}, as this does not contribute to the optimization.

\subsection{Ground Plane Segmentation}
To segment the ground plane we exploit the fact that the ground plane is below the object and that it is visible on the image near the object bounding box. To do so, first we crop the recovered Single-Image depth map -- to fit the ROI active tracks, backproject it to an organized point cloud and rotate it using the camera pose to align it with gravity. We then estimate the object coarse height by averaging the height values of all points within the target bounding box. \par

The gravity-aligned point cloud is then segmented using the grid illustrated in Fig. \ref{fig:fig2}.b. For each tile that is not contained in the target bounding box we fit plane to the respective points. Tiles that cross the bounding box are dilated outward forming a \textit{seeding region} where we sample tiles for region growing. Specifically, first we remove tiles from the \textit{seeding region} that have an average point height higher than the object's height estimate. We then select as a seed the tile with minimum Mean Squared Error (MSE) given the plane fitting. Tile-wise Region Growing is performed on the white tiles shown in Fig. \ref{fig:fig2}.b, and alternated with the seeding to cover possible split ground plane as shown in Fig. \ref{fig:fig3}. The Region Growing algorithm is inspired by \cite{proencca2018fast}, it uses a 4-neighbour search and proceeds as follows: A tile neighbour $c$ of a current seed $s$ is added to the growing region if: (i) the dot product of their plane normals exceeds a threshold, (ii) the distance from the centroid in $c$ to the plane in $s$ is lower than $d \sin \alpha$ where $d$ is the Euclidean distance between centroids and $\alpha$ is set to 3$^\circ$ and (iii) the MSE of the plane in $c$ is lower than 0.001.

As shown in Fig. \ref{fig:fig3}, the resulting mask from tile-wise Region growing will be coarse thus as a final step we refine the tiles on the border by checking for all their pixels if the distances to the plane solution given by all points in the grown region is less than 9 times the plane MSE.
Once the ground mask is obtained, we can now check which ROI features are inside it and use these for plane fitting.

\begin{figure*}[tb]
	\centering
	\begin{tabular}{@{}c@{\hspace{3pt}}c@{\hspace{3pt}}c@{}}
    \includegraphics[width=0.35\textwidth]{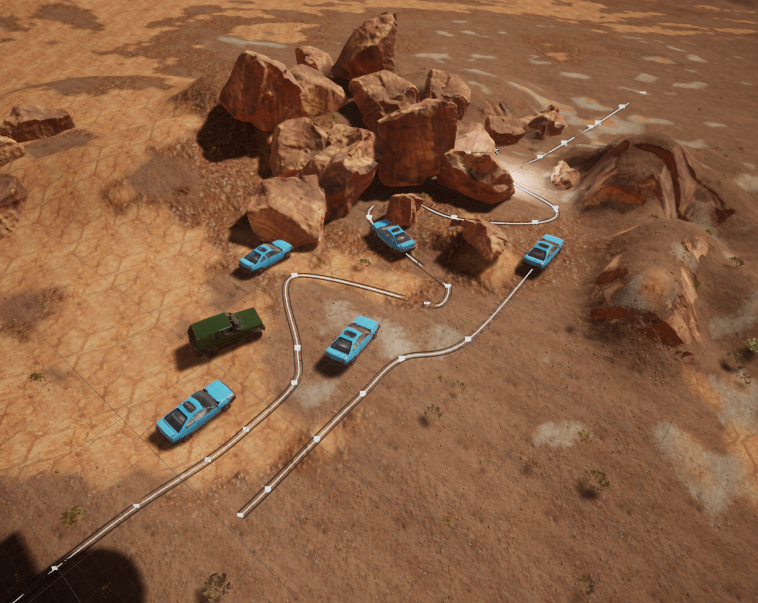} &
    \includegraphics[width=0.3\textwidth,trim={0cm 0cm 0cm 1cm},clip]{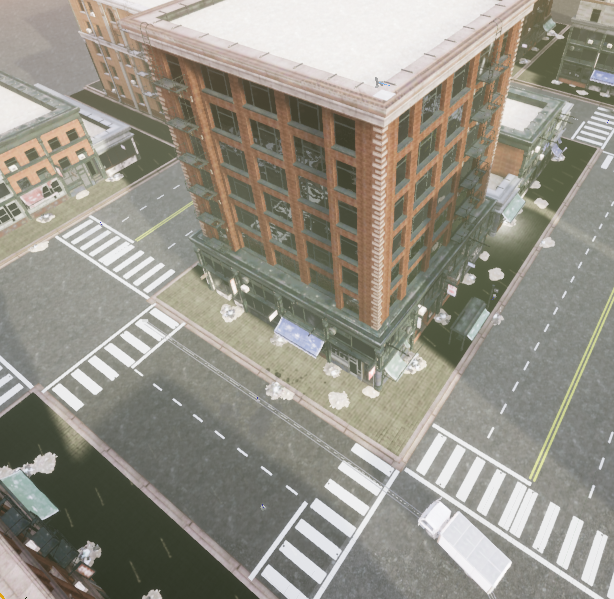} &
    \includegraphics[width=0.34\textwidth,trim={0cm 1cm 0cm 5.45cm},clip]{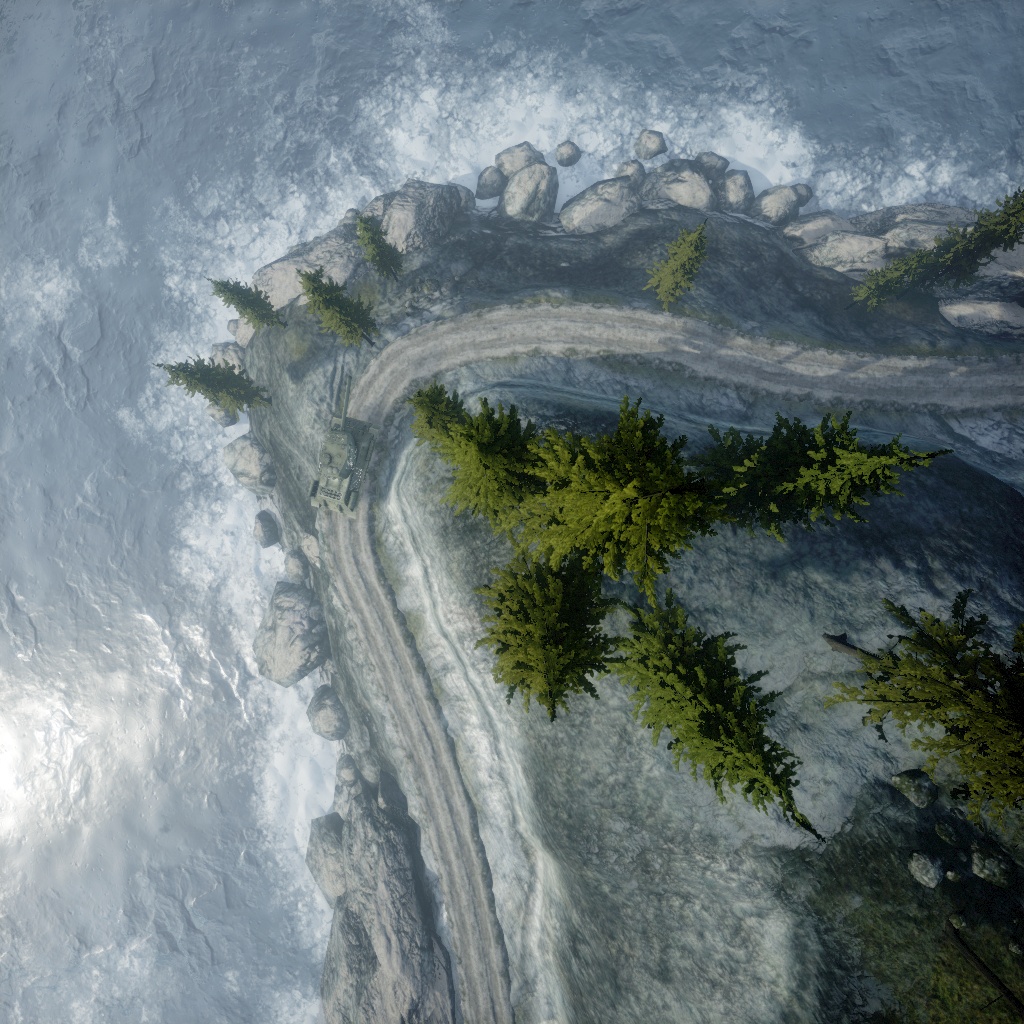}
    \end{tabular}
	\caption{Scenes used in this work. \textit{Left}: Rocky desert with several cars identical to the target car which moves through the rock field following a spline (shown as a white line). The drone flight is set to follow the target. \textit{Center}: City. The drone takes off from the building's rooftop and flies along its edge to follow the white truck. \textit{Right}: Mountain with icy lake where a tank moves along the dirt-road}
	\label{fig:fig5}
\end{figure*}

\subsection{Multi-RANSAC plane fitting}
\label{sec:RANSAC_plane}

Given a set of 3D points around the target, we can estimate the ground plane parameters by fitting a plane to these. We use a RANSAC loop with MSAC criteria to reject depth outliers and also because there can be multiple planes within the point cloud. Moreover, we have experienced that the plane that maximizes the number of inliers is often a plane traversing multiple surfaces. Thus, we use a Multiple model RANSAC scheme that finds the $N$ largest planes that have significantly different plane parameters between them, i.e. angle between normals must exceed $15^\circ$ and they should not share more than 80 $\%$ of their inliers. Plane solutions that have less inliers than a fraction of the number of inliers of the largest plane solution are discarded. Finally, we select the flattest plane from this set of plane solutions based on the plane normals. Another advantage of Multi-RANSAC is that we get a depth uncertainty estimate by ray-casting the multiple plane solutions and fit a Gaussian to the depth estimates.

\subsection{Temporal fusion}

The ground mask can be under-segmented or the ground plane can be temporally covered (e.g. trees or mountains) resulting in few ROI points and in turn noisy plane estimates.
Therefore, we found beneficial to use planes and depth estimates from past frames. To do so, we accumulate in a buffer the point inliers given by the plane solution -- estimated in the last section. If the number of points in the buffer exceeds a certain number we remove the oldest points. Using these accumulated points we run the Multi-RANSAC plane fitting once more and store the plane solution for the next frames. This way if the ground mask fails or number of ROI is too low in the next frames we reuse this plane to raycast target depth. Additionally, since the ground mask can also be over-segmented and lead to a wrong estimated plane we gate the incoming points by checking if they are less than a given distance from the past plane solution. This distance threshold is set based on MSE of the plane that fits all accumulated points. This gating enforces temporal consistency assuming that the ground plane does not change rapidly.

\section{Experiments}
\label{sec:experiments}
We implemented two versions of TRADE. For offline evaluation, we used the full TRADE pipeline, depicted in Fig. \ref{fig:fig1}, whereas for real-time applications we deployed a lighter TRADE version on a Voxl 2 board. These are described respectively in Section \ref{sec:off_implementation_details} and \ref{sec:real_implementation_details}.

\subsection{Synthetic Datasets and Evaluation Methodology}
\label{sec:datasets}
There are many tracking datasets \cite{kristan2020eighth} but to our knowledge none includes ground-truth depth, camera poses and IMU. Therefore to evaluate object 3D localization and tracking we created 3 scenes using Unreal Engine 4, shown in Fig. \ref{fig:fig5}: \textit{Rocky desert}, \textit{City} and \textit{Mountain}. Each scene has one object target, which is moved along a Spline. Using the Airsim \cite{shah2018airsim} plugin we set drone flights to follow the target and collected IMU at 200 Hz and 1024$\times$1024 px images at 30 fps using a downward-looking camera with 110$^\circ$ of FOV. One sequence was recorded per scene, and for each of these sequences we generated 5 noisy sequences by adding Shot noise to the images and IMU noise based on the MPU9250 noise model. 
Camera pose was obtained by running xVIO \cite{delaune2021range} on each noisy sequence. Additionally for evaluating tracking we randomly shifted the initial tracking frame within 10 frames and we initialized the tracker with the respective ground-truth bounding box coordinates perturbed by an offset randomized within an interval of 5 pixels. \par
For the \textit{Rocky desert} the drone flew horizontally 90 m at 16 m of altitude above ground, for the \textit{City} the drone flew horizontally 20 m at 32 m of altitude above ground and for the mountain the drone flew horizontally 82 m at 30 m of altitude above ground with a 65$^\circ$ yaw turn after 50 m.

\subsection{Offline Implementation Details}
\label{sec:off_implementation_details}

We used DIMP \cite{bhat2019learning} as our object tracker in our full TRADE implementation. In terms of ground feature tracking, the ROI structure for detecting features was set by adding margins to the target bounding box. These margins were 2.0 times the minimum side of the bounding box for both \textit{City} and \textit{Mountain} and 1.0 for the \textit{Rocky desert} where the ground is more textured. For each ROI tile, from the 3$\times$3 grid, we set the maximum number of features per tile: 25. For temporal fusion we set the maximum number of points to 1000. For multi-RANSAC plane fitting we set the inlier threshold to 0.5 m and use a total of 4 solutions with at least half the number of inliers of the top solution. For ground plane segmentation in full TRADE, the point cloud is split into 20$\times$20 tiles and for its region growing, we use a total of 8 seeds. Even though, the plane segmentation is not optimized, the algorithm is an adaptation of \cite{proencca2018fast} which can run at 300 fps on a single CPU thread.

\subsection{Real-time Implementation and Target Following Details}
\label{sec:real_implementation_details}

A lighter TRADE version was deployed on a Voxl 2 CPU. For this, we used STARK \cite{yan2021learning} instead of DIMP. STARK runs at about 5-10 fps on this CPU using OnnxRuntime while sharing the CPU cores with VIO \cite{delaune2021range}. Thus, the object localization module (shown in Fig. \ref{fig:fig1}) was moved to a separate C++ thread to keep up with the camera frame-rate. This thread performs only ROI Feature Tracking and robust plane fitting with temporal fusion to estimate and update the ground plane model. The single-image depth and ground plane mask modules were not included in this light-version as the scenes in our real-world experiments are approximately planar. \par
As shown in Fig. \ref{fig:fig0} we used a FLIR Boson thermal camera mounted with a 45$^\circ$ pitch on a quadcopter.
For target following, waypoints are continuously sent to the PX4 flight controller based on the 3D object filtered location. This is given by the Kalman Filter that is used for trajectory prediction (in Section \ref{sec:traj_prediction}). This results in a smooth drone path despite unstable bounding boxes. The waypoints are send to keep the same drone altitude and yaw while a horizontal offset is added to keep the object centered in the FOV.

\begin{figure}[t]
	\centering
	{\small Rocky desert}
    \includegraphics[width=\textwidth,trim={1.9cm 0.6cm 2.5cm 0.7cm},clip]{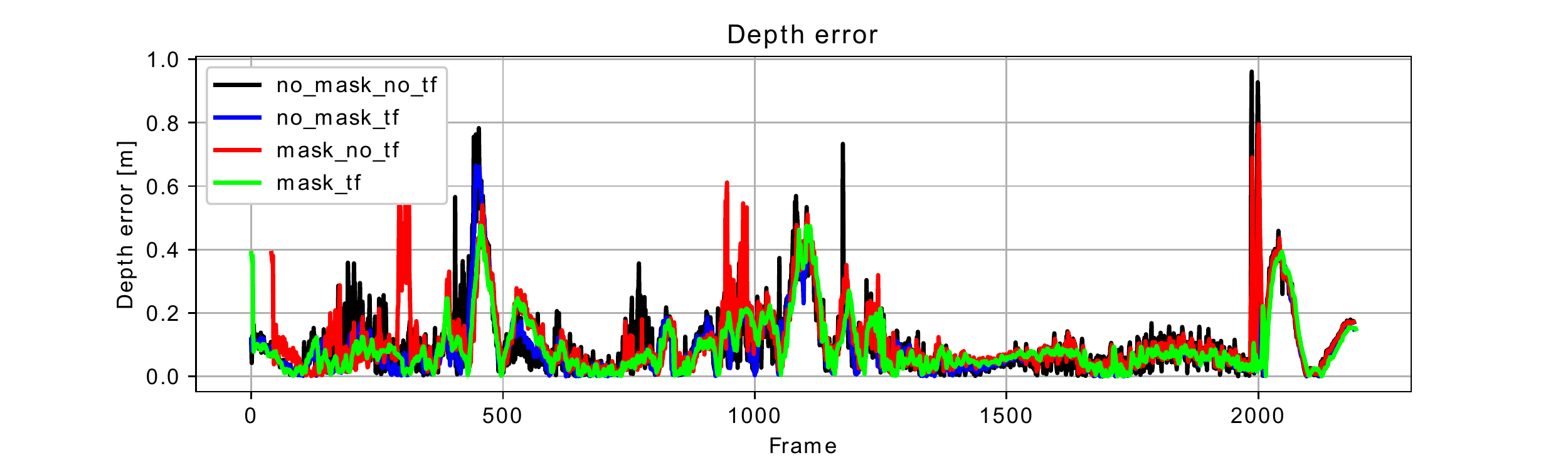}
    {\small Mountain}
    \includegraphics[width=\textwidth,trim={1cm 0.6cm 2cm 0.8cm},clip]{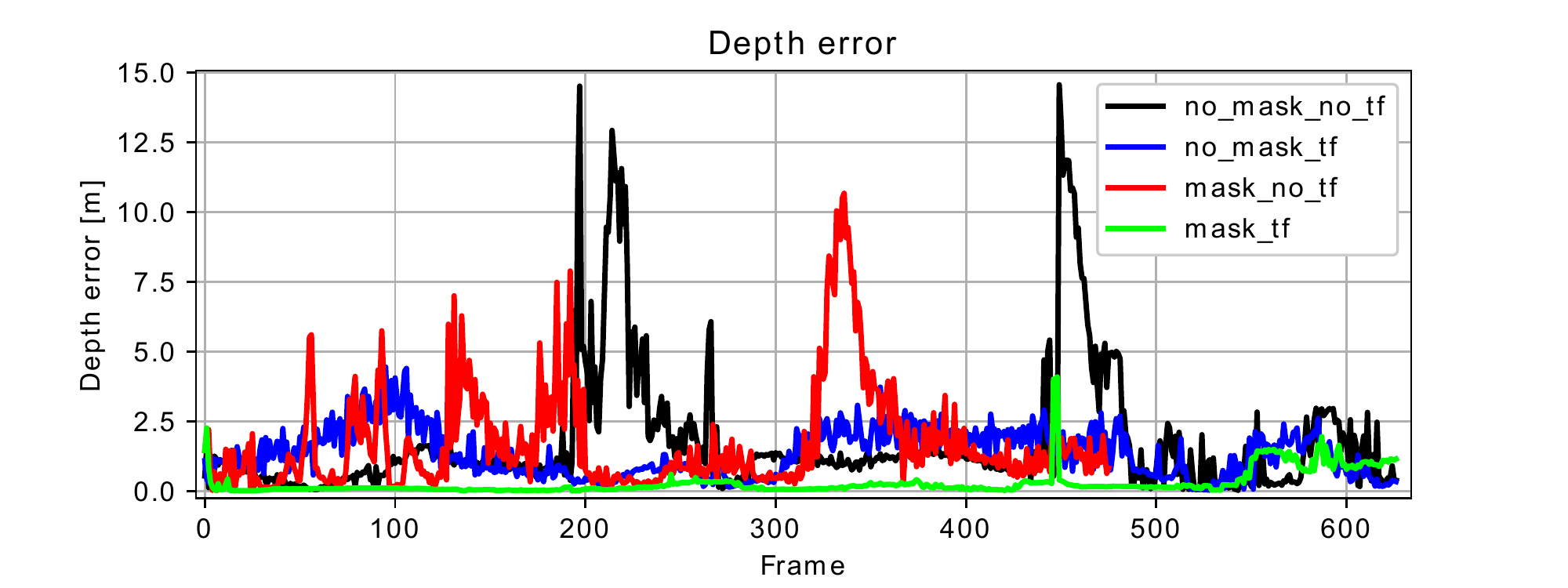} 
    {\small City}
    \includegraphics[width=\textwidth,trim={1cm 0cm 2cm 0.8cm},clip]{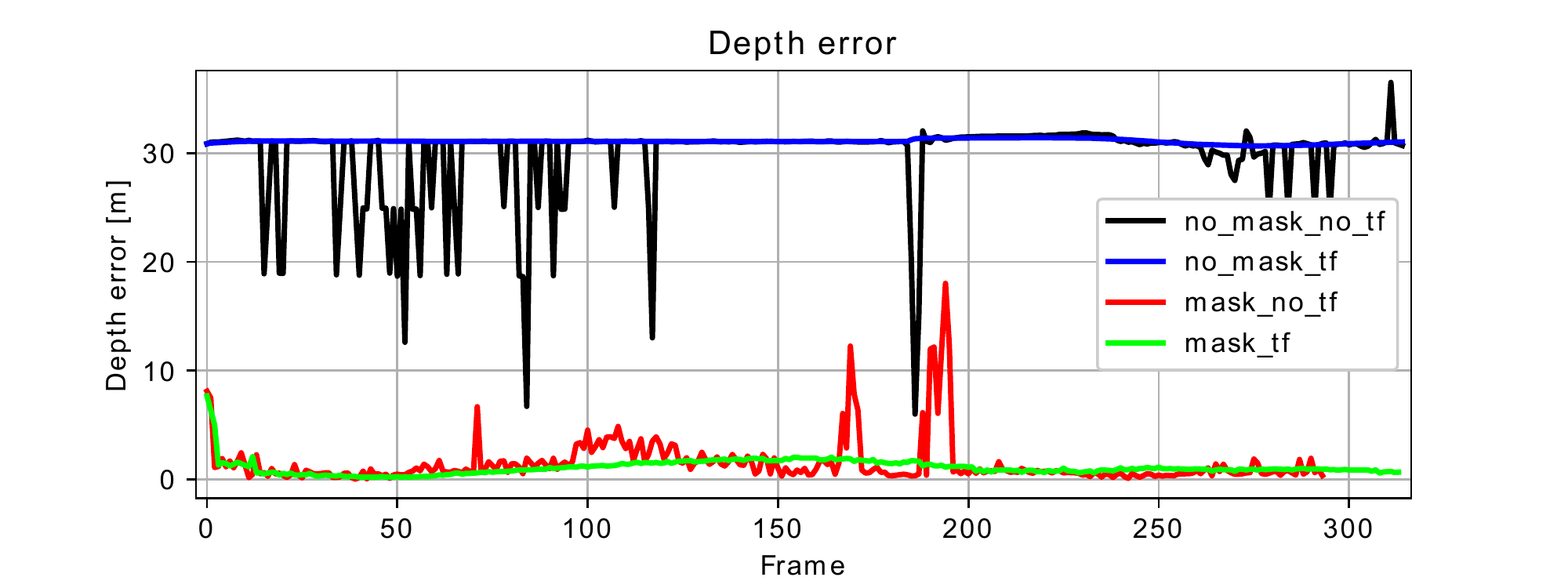} 
	\caption{Depth error per frame averaged over 5 runs, corresponding to Table \ref{tab:tab1}. \textit{no\_mask\_no\_tf} stands for disabled ground mask and temporal fusion whereas \textit{mask\_tf} stands for full TRADE.}
	\label{fig:fig6}
\end{figure}

\subsection{Results: Depth Estimation}

To evaluate depth estimation, we disabled tracking and used the ground truth bounding boxes to remove the influence of tracking errors.
Each sequence was repeated 5 times due to RANSAC stochastic behaviour. Table \ref{tab:tab1} and \ref{tab:tab2} show the RMSE of depth estimation for all frames in the 5 runs. Fig. \ref{fig:fig6} shows the average depth error per frame.

First, Table \ref{tab:tab1} shows the results for ground depth estimation by ray-casting the ground truth bounding box center. The ground truth depth is given by the terrain surface. ROI feature depth estimation was performed using either VIO or ground truth pose to separate the effect of camera pose errors. The results are overall better on the \textit{Rocky desert} since the flight altitude is lower and the ground is more consistently textured. Temporal fusion improves overall the results especially when combined with the ground plane mask. On the \textit{Rocky desert}, our multi-RANSAC plane fitting is already successful in rejecting depth measurements from the rocks. Whereas on the \textit{City} the mask is necessary to reject measurements from the building which is about 30 m tall. On the \textit{Mountain}, combining the mask with temporal fusion also leads to more robustness, as shown in Fig. \ref{fig:fig6}. Results using VIO poses are significantly worse than using the ground-truth pose, indicating that this filter-based VIO needs to be improved or fine-tuned for such high-altitude flights.

Instead of ground plane estimates, Table \ref{tab:tab2} shows the results of estimating the object's depth simply by sampling the Single-Image depth map after affine transformation, which we estimate as described in Section \ref{sec:SIdepth} with and without RANSAC.
Here we use the ground-truth depth of the target to directly evaluate the accuracy. We have found the performance to be inconsistent throughout the sequence. While using RANSAC improves the RMSE on the \textit{Rocky desert}, on the \textit{City} it makes it worse by limiting the RANSAC samples to the building. This can be improved by using a more sophisticated sampling scheme. But overall the performance is far from the ground plane estimates.

\begin{table}
\small{
\begin{tabular}{|ll|lll|}
\hline
\multicolumn{1}{|c}{\begin{tabular}[c]{@{}c@{}}Temporal\\ fusion\end{tabular}} & \multicolumn{1}{c|}{\begin{tabular}[c]{@{}c@{}}Ground \\ mask\end{tabular}} & \multicolumn{1}{l}{\begin{tabular}[c]{@{}l@{}}Rocky \\ desert\end{tabular}} & \multicolumn{1}{l}{City} & \multicolumn{1}{l|}{Mountain} \\ \hline \hline
\ding{55} & \ding{55} & 0.16 (1.64) & 30.4 (30.3) & 3.05 (5.20) \\
\ding{55} & \checkmark & 0.17 (1.55) & 3.74 (4.07) & 1.83 (5.04) \\
\checkmark& \ding{55} & \textbf{0.13} (1.39) & 31.1 (31.0) & 3.00 (3.83) \\
\checkmark & \checkmark & \textbf{0.13 (1.26)} & \textbf{1.46 (3.5)} & \textbf{0.55 (2.98)} \\ \hline
\end{tabular}
}
\caption{RMSE of ground depth estimates [m] using either ground-truth camera pose or VIO \cite{delaune2021range} in brackets. Ground segmentation and temporal fusion were switched on-off.}
\label{tab:tab1}
\end{table}


\begin{table}
\small{
\begin{tabular}{|l|lll|}
\hline
RANSAC & Rocky desert & City & Mountain \\
\hline \hline
\ding{55}  & 4.69 & 14.77 & 4.54 \\
\checkmark  & 1.66 & 26.72 & 4.60 \\
\hline
\end{tabular}}
\caption{RMSE of depth estimates [m] using Single Image depth. Ground-truth camera pose was used for feature depth estimation.}
\label{tab:tab2}
\end{table}

\begin{figure}
	\centering
	{\small Rocky desert}
    \includegraphics[width=\textwidth,trim={1.5cm 0.6cm 2.4cm 0.7cm},clip]{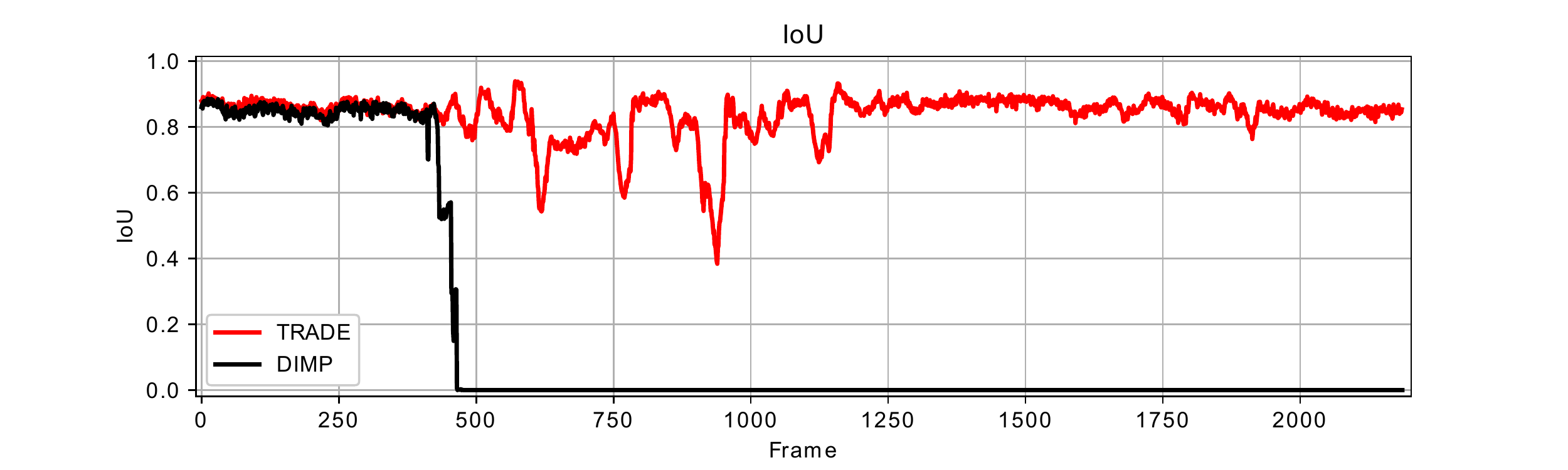}
    {\small Mountain}
    \includegraphics[width=\textwidth,trim={1.5cm 0.6cm 2.4cm 0.7cm},clip]{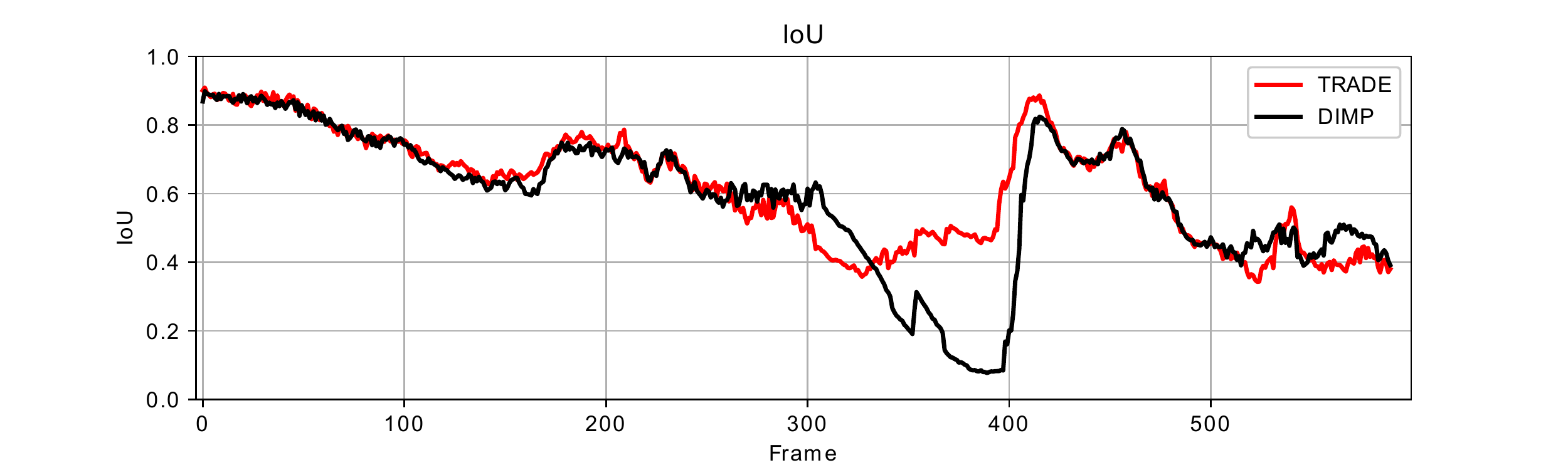}
    {\small City}
    \includegraphics[width=\textwidth,trim={1.5cm 0cm 2.4cm 0.7cm},clip]{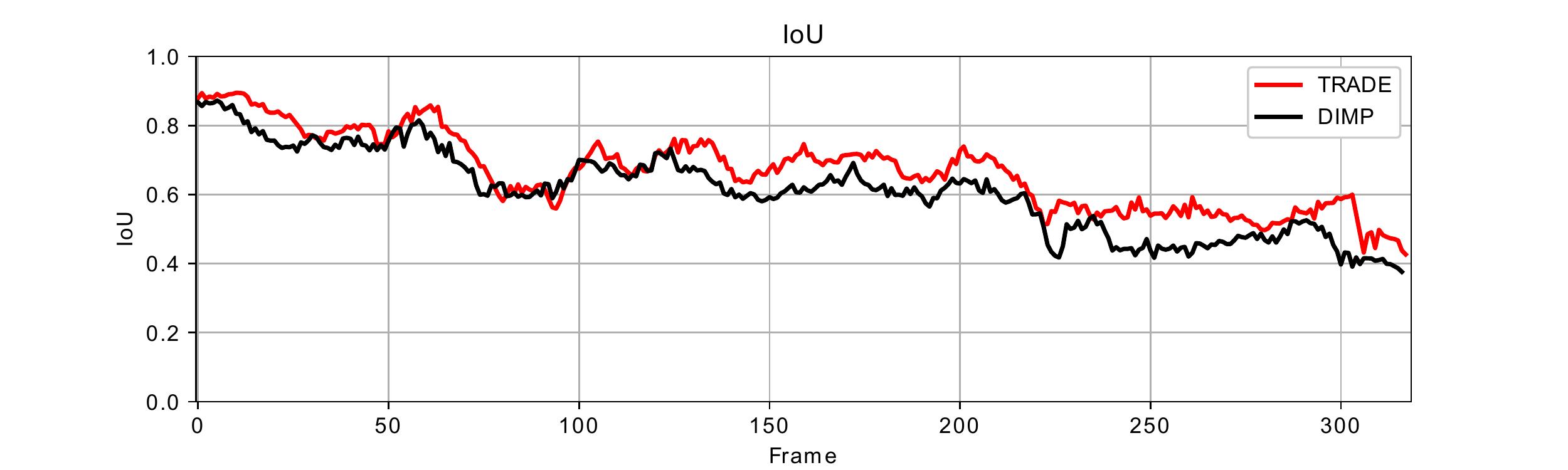} 
	\caption{Tracking accuracy as bounding box IoU per frame averaged over 10 runs. Camera pose was estimated using VIO \cite{delaune2021range}}
	\label{fig:fig7}
\end{figure}

\subsection{Results: Object Tracking}

\begin{figure}
	\centering
	\begin{tabular}{@{}c@{\hspace{3pt}}c@{}}
	\includegraphics[width=0.5\textwidth,trim={0cm 0cm 0cm 1.5cm},clip]{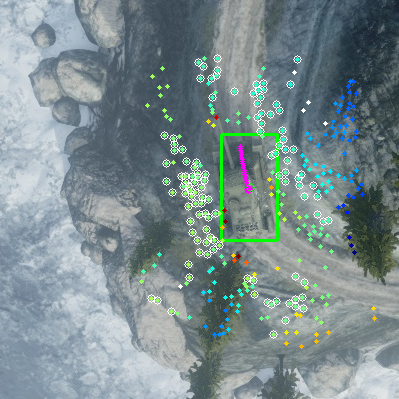} &
    \includegraphics[width=0.5\textwidth,trim={0cm 0cm 0.4cm 0.5cm},clip]{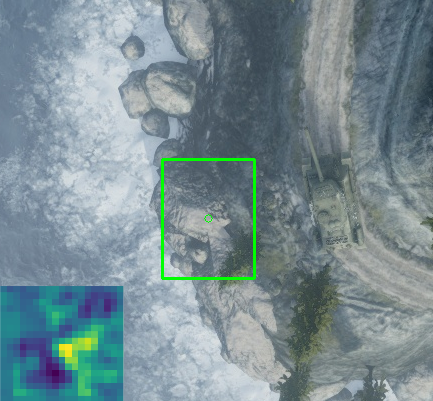}
    \end{tabular}
	\caption{Examples of challenging tracking. \textit{Left:} TRADE using predicted trajectory and ground feature depth estimates (color-coded based on depth). Features with a white circle are the plane fitting inliers. \textit{Right:} On the same frame
	DIMP \cite{bhat2019learning} performs \textit{target switching} due to multi-peak tracker's response shown at the corner.}
	\label{fig:fig8}
\end{figure}

Object tracking performance is evaluated between TRADE and DIMP on 10 Monte Carlo samples per sequence. Results are shown in Fig. \ref{fig:fig7} as the Intersection-over-Union (IoU) between the ground-truth and estimated bounding boxes. On the \textit{Rocky desert}, DIMP fails all 10 runs due to the \textit{target switching} shown in Fig. \ref{fig:fig0}. There's also partial occlusion, which causes the DCF to loose temporally tracking. This appears as the lowest average IoU for TRADE in Fig. \ref{fig:fig7}. On the \textit{Mountain}, DIMP also suffers from \textit{target switching}, shown in Fig. \ref{fig:fig8}. whereas TRADE is constrained by the trajectory seen as magenta on Fig. \ref{fig:fig8}. On the \textit{City}, the differences are not that significant. For more details check our supplementary video.

\section{Conclusion and Limitations}

The combination of all 3d localization components in TRADE shows improved robustness to complex terrain. We also demonstrate a practical approach to improve tracking robustness by incorporating the trajectory predictions. However a few concerns remain. The tracking performance is sensitive to the fixed process and measurement noise of the Kalman filter. A more adaptive parameterization as in \cite{coskun2017long} is desirable. Using a dedicated single-image depth estimation process introduces a significant computational cost. This can be alleviated by using only when necessary, e.g., during initialization. We demonstrated a simple but effective target-following operation using TRADE onboard. This can be further improved by taking into account motion planning, control and the camera pitch and FOV.

\section{Acknowledgments}

The research was funded by the Combat Capabilities Development Command Soldier Center and Army Research Laboratory. The research was carried out at the Jet Propulsion Laboratory, California Institute of Technology, under a contract with the National Aeronautics and Space Administration (80NM0018D0004). \copyright 2022 California Institute of Technology.

\bibliographystyle{ieeetr} 
{\footnotesize
\bibliography{icra_ref}
}

\end{document}